%% file: main.tex
\RequirePackage{silence}
\documentclass[runningheads]{llncs}

\usepackage{xcolor}
\usepackage{graphicx}
\graphicspath{{/}{fig/}}
\usepackage{cite}
\usepackage{tikz}
\usepackage{array}
\usepackage{textcomp}
\usepackage{xcolor}
\usepackage{multirow}
\usepackage{booktabs}
\usepackage{float}
\usepackage{caption}

\usepackage{pgfplots}
\pgfplotsset{compat=1.7}
\usepgfplotslibrary{groupplots}

\usepackage[font=small]{caption}
\usepackage{subcaption}

\usepackage{multirow,tabularx}
\usepackage{hyperref}
\usepackage{flushend}
\usepackage{algorithmic}

\usepackage[vlined, ruled, shortend]{algorithm2e}

\usepackage{pgfmath}
\usetikzlibrary{decorations.text, arrows.meta,calc,shadows.blur,shadings}

\usepackage{float}
\usepackage{tikz}

\newlength\figureheight
\newlength\figurewidth
\SetAlCapNameFnt{\normalsize}
\SetAlCapFnt{\normalsize}

\begin{document}
\title{Is Alice Really in Wonderland?\\UWB-Based Proof of Location for UAVs with Hyperledger Fabric Blockchain}
\author{
    Lei Fu\inst{1} \and
    Paola Torrico Mor\'on\inst{1} \and
    Jorge Peña Queralta\inst{1} \and
    David H\"astbacka\inst{2} \and
    Harry Edelman\inst{3} \and
    Tomi Westerlund\inst{1}
}%
\authorrunning{Lei Fu et al.}
\titlerunning{UWB-Based Proof of Location for UAVs with Blockchain Identities}
\institute{
        Turku Intelligent and Embedded Robotic Systems, University of Turku, Finland\\ 
        \email{\{leifu, pctomo, jopequ, tovewe\}@utu.fi} |
        \url{https://tiers.utu.fi}
    \and
        Department of Engineering and Business, 
        Turku Univesity of Applied Sciences, Finland |
        \email{harry.edelman@turkuamk.fi} \\
    \and
        Computing Sciences, Faculty of Information Technology and Communication Sciences, Tampere University, Finland | \email{david.hastbacka@tuni.fi} \\
    }
\maketitle
%
%
%
%%%%%%%%%%%%%%%%%%%%%%%%%%%%%%%%%%%%%%%%%%%%%%
%%                                          %%
%%                SECTIONS                  %%
%%                                          %%
%%%%%%%%%%%%%%%%%%%%%%%%%%%%%%%%%%%%%%%%%%%%%%
\input{sections/00_Abstract}

\input{sections/01_Introduction}

\input{sections/02_RelatedWork}

\input{sections/03_Solution}

\input{sections/04_Methodology}

\input{sections/05_Conclusion}

\section*{Acknowledgment}

This research work is supported by the Academy of Finland's AeroPolis project (Grant No. 348480, 348481) and by the R3Swarms project funded by the Secure Systems Research Center (SSRC), Technology Innovation Institute (TII).

%%%%%%%%%%%%%%%%%%%%%%%%%%%%%%%%%%%%%%%%%%%%%%
%%                                          %%
%%              BIBLIOGRAPHY                %%
%%                                          %%
%%%%%%%%%%%%%%%%%%%%%%%%%%%%%%%%%%%%%%%%%%%%%%

% \nocite{*}
\bibliographystyle{splncs04}
\bibliography{bibliography}

\end{document}

%% file: sections/00_Abstract.tex
\begin{abstract}

    Remote identification of Unmanned Aerial Vehicles (UAVs) is becoming increasingly important since more UAVs are being widely used for different needs in urban areas. For example, in the US and in the EU, identification and position broadcasting is already a requirement for the use of drones. However, the current solutions do not validate the position of the UAV but its identity, while trusting the given position. Therefore, a more advanced solution enabling the proof of location is needed to avoid spoofing. We propose the combination of a permissioned blockchain managed by public authorities together with UWB-based communication to approach this. Specifically, we leverage the identity management tools from Hyperledger Fabric, an open-source permissioned blockchain framework, and ultra-wideband (UWB) ranging, leading to situated communication (i.e., simultaneous communication and localization). This approach allows us to prove both the UAV identity and also the location it broadcasts through interaction with ground infrastructure in known locations. Our initial experiments show that the proposed approach is viable and UWB transceivers can be used for UAVs to validate both their identity and position with ground infrastructure deployed in known locations.
    
    \keywords{
        UAV \and 
        Remote Identification \and
        Proof of Location \and 
        Blockchain \and 
        Ultra-Wideband \and 
        UAV \and 
        Unmanned Aerial Systems
    }

\end{abstract}

%% file: sections/01_Introduction.tex
\section{Introduction}
\label{sec:intro}

Unmanned aerial vehicles (UAVs) are becoming increasingly popular for a wide range of applications, from aerial surveying to UAV delivery services~\cite{bithas2019survey, osco2021review, mohsan2022towards}. However, the integration of UAVs into the airspace poses significant challenges, particularly in terms of ensuring the safety and security of other airspace users and the public~\cite{outay2020applications, kakaletsis2021computer, xiangmin2020survey}. One critical challenge is the need for a reliable and trustworthy proof of location system for UAVs. Therefore, direct remote identification, the process of identifying an UAV from a remote location without an access to the UAV,  has been introduced~\cite{belwafi2022unmanned}. It is a crucial aspect of UAV regulation and safety as it enables authorities to identify the operator and the UAV in real-time, which is important for ensuring compliance with regulations and preventing potential security threats~\cite{stocker2017review, khan2022detection}. In Europe, the U-space is evolving as a regulatory framework for the safe and efficient operation of UAVs, with the more standard \textit{unmanned aircraft systems} (UAS) name in the regulations. The U-space includes a range of measurements such as the registration and direct remote identification and real-time tracking of the UAS. Same in the United States, the Federal Aviation Administration (FAA) has issued regulations that require UAVs to have remote identification capabilities~\cite{batuwangala2018regulatory}. 

\begin{figure}[t]
    \centering
    \includegraphics[width=\textwidth]{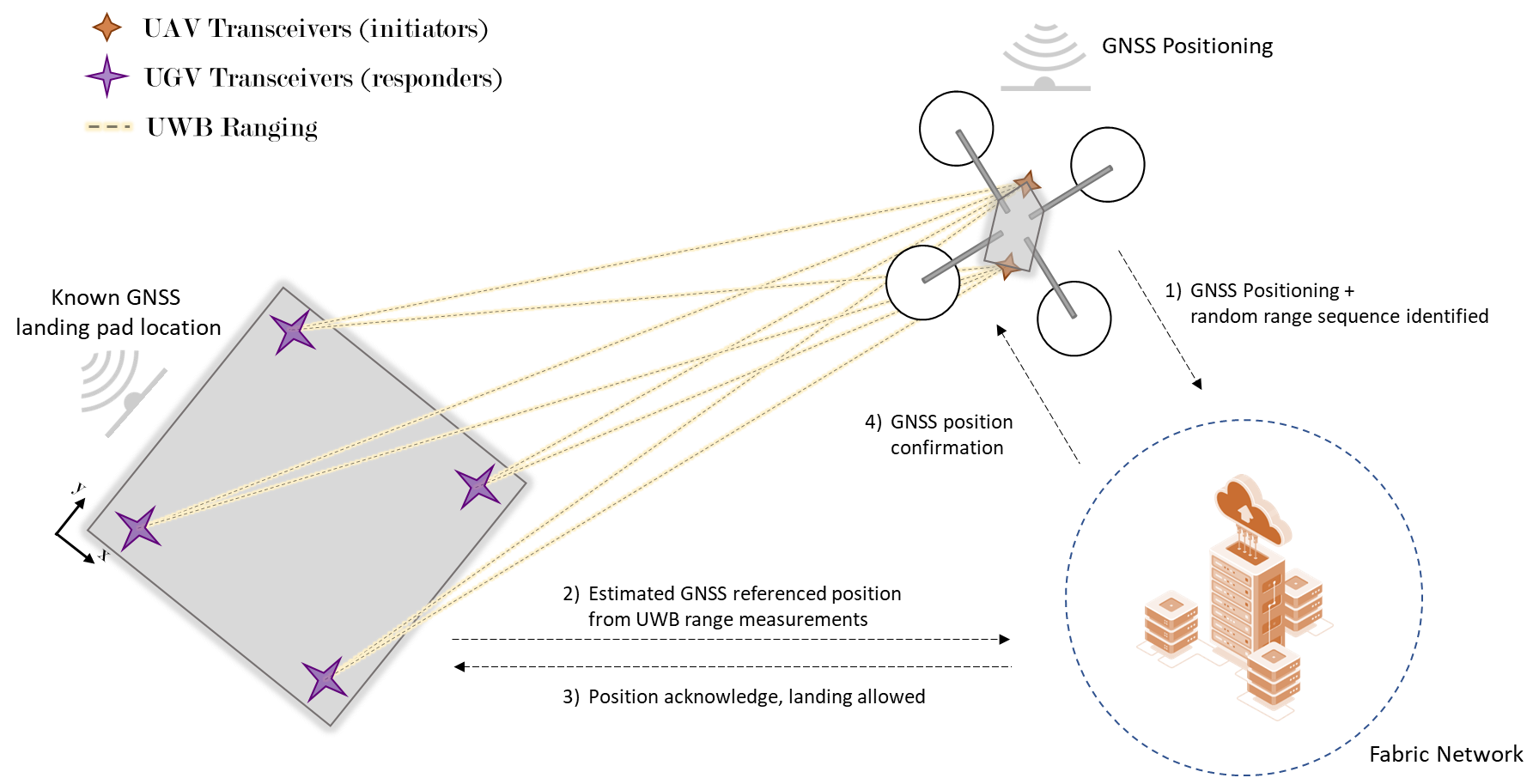}
    \caption{Illustration of the proposed remote identification process.}
    \label{fig:remote_id}
\end{figure}

However, the risk of spoofing in global navigation satellite system (GNSS)can be a potential problem for direct remote identification of UAVs~\cite{sorbelli2020uavs, chamola2021comprehensive, meng2022survey}, which is the deliberate manipulation of signals to deceive the remote identification system~\cite{tang2021review}. As for proof of location (PoL), while it could be a useful tool to verify the location of a UAV, it is not always secure. Even with trust, there is a risk of errors or intentional manipulation of location data. For example, GNSS signals can be jammed or spoofed with incorrect location data as well as environment conditions such as interference could affect the accuracy of location data. In this chapter, we propose a novel PoL solution for creating a trustable and secure air-ground interaction ecosystem for short-distance UAV operation. The solution combines Ultra-Wideband (UWB) technology with decentralized ledger technology (DLT), such as Hyperledger Fabric blockchain, to generate highly accurate and reliable location data for UAVs, which can be securely stored and verified on a tamper-proof and decentralized blockchain. The concept is illustrated in Fig.~\ref{fig:remote_id}.

The rest of this chapter is structured as follows. In Section\,2, we introduce background concepts in the areas of Proof of Location, UWB technology and the Hyperledger Fabric blockchain framework. Section\,3 then describes related works in the area of proof of location and remote identification for UAVs.

\section{Background}
\label{sec:background}

Through this section, we briefly introduce the key technologies behind the UWB-based Proof of Location approach for UAVs with Hyperledger Fabric blockchain.

\subsubsection{Proof of Location} (PoL) is a technology that enables the verification and broadcasting of a device's physical location coordinates to a blockchain network~\cite{waters2003secure, javali2016alice, amoretti2018blockchain, zafar2020location}. It enables other devices to rely on the location data without having to trust the broadcasting device. The technology uses cryptographic methods to prove the authenticity and accuracy of the location data, making it a secure and trustworthy source of location information. In the field of UAVs, PoL can be a valuable tool for ensuring accountability and security in critical applications, such as UAV deliveries and autonomous flying. By integrating PoL with blockchain identities, UAVs can be better managed and monitored. This integration can provide enhanced security, reliability, and accountability by ensuring that the UAV is at a specific location when a critical action is performed, such as landing or taking off. It also has potential from the perspective of access control and secure fleet management.

\subsubsection{Ultra-Wideband} (UWB) technology is one of the most accurate wireless ranging solutions that can enable PoL implementations~\cite{zhang2006secure, wilson2007channel, singh2019uwb}. UWB uses radio frequency signals to determine the location of a device with high precision and broadcast the encrypted data and signature to the other nodes, providing a reliable source of real-time location data and ensuring UAVs' verification~\cite{queralta2022viouwb} By segmenting the data into smaller parts, encrypting, digitally signing, and then transmitting each part as a separate message, the accuracy and trustworthiness of location data can be increased. Even if an attacker intercepts some of the segments, they will be unable to use the information without having access to all the parts, providing an added layer of security. UWB technology has a range of applications in addition to PoL, including indoor positioning, asset tracking, and even remote control of devices~\cite{xianjia2021applications}. Its high accuracy and low power consumption make it an attractive choice for a wide range of IoT and wireless communication applications. Air-ground coordination and localization systems based on UWB have already been proposed in the literature~\cite{xianjia2021cooperative}.

\subsubsection{Hyperledger Fabric} is a blockchain platform that offers various features beneficial to distributed robotic systems, including secure identity management and certificate generation for participant identification, permissioned networks for enhanced data security, high performance capabilities for real-time robotic data processing, low-latency transaction confirmation for real-time consensus, and partitioned data channels enabling transaction privacy and confidentiality~\cite{salimi2022towards, lei2023event}. Moreover, these features can integrate with the pub/sub system of ROS, providing a reliable and scalable system for distributed robotic systems, making it a potential candidate for use in PoL mechanisms of UAVs and similar applications where secure data exchange and privacy preservation are critical factors.

In this scenario, the integration of UWB and blockchain technology, such as Hyperledger Fabric, can be an effective solution for PoL mechanisms in the field of UAVs. As we have demonstrated in previous papers integrating aerial and ground robot systems with UWB and Hyperledger Fabric, UWB provides real-time location information and, with the latest standards, also  encrypted data transmission with high accuracy and low latency, while the blockchain acts as a secure ledger to store and verify the authenticity of the location information~\cite{salimi2022secure}. This combination makes it a potential solution for increasing the accuracy of UAV location and has wide-ranging applications in industries such as delivery, logistics, and aerospace. By leveraging these technologies, PoL can be extended to improve the efficiency, security, and transparency of location-based applications.

%% file: sections/02_RelatedWork.tex
\section{Related work}
\label{sec:related}

The concept of Proof of Location (PoL) has been proposed as a means of validating user locations in various studies. Initially, Brent Waters and Felten Edward introduced the concept of PoL in a centralized setting, where a trusted location manager signs the proof based on round-trip communication measurements with the prover~\cite{waters2003secure}. The concept of proof of location was first proposed by Brent Waters and Felten Edward in a centralized setting, where a trusted location manager signs the proof based on the round-trip communication measurement with the prover. 
Recently, new approaches have been proposed for generating PoL to ensure the location of UAVs as well as other mobile assets within the more general domain of the Internet of Things (IoT).

For instance, Khan et al. introduced the concept of witness-oriented endorsements and a collusion-resistant protocol for generating asserted location proofs and identifying the vulnerabilities that may arise in non-federated environments~\cite{khan2014and}. In another work, Chitra et al. proposed a new protocol for generating and verifying secure location proof~\cite{javali2016alice}. The protocol employs an information theoretically secure fuzzy vault scheme and unique spatial-temporal wireless channel characteristics as well as offers a cryptographic security. In a more recent work, and incorporating distributed ledger technologies, Amoretti et al. presented a novel method of blockchain-based PoL in 2018~\cite{amoretti2018blockchain}. The approach consisted on producing digital certificates that attest the presence at a specific geographic location at a certain point in time by using a decentralized peer-to-peer scheme that ensures location trustworthiness and user privacy.

In addition to these, there has been growing interest in the development of remote identification systems for UAVs in recent years as a means of improving safety and security in the airspace. Various approaches have been proposed, including radio frequency (RF) identification, and other sensor-based methods. In terms or remote detection and identification of UAVs, Mohammad et al. in~\cite{al2019rf} developed a comprehensive database containing RF signals emitted by UAVs in different flight modes. Based on this database, novel algorithms have been proposed to detect and identify UAVs, as well as determine their type and flight modes.

In terms of blockchain integration for remote identification of drones, the authors in~\cite{hashem2021secure} introduced a new Drone Remote Identification Protocol (DRIP) using Hyperledger Iroha framework and its applications to identify UAVs securely. The main idea of this work is to register new UAVs and store the identification to the blockchain network and validate the received data by the public key and certificates. However, the system deals with identification alone and does not consider the proof of location. Therefore, there is a gap in the literature in the area of remote direct identification with proof of location for UAVs.

%% file: sections/03_Solution.tex
\section{Proposed System Architecture}

We propose the integration of UWB ranging for remote UAV IDs to enable proof of location when the UAVs interact with infrastructure, such as landing pads in known locations. Both the UAV and the infrastructure must be part of a common Fabric blockchain.

In this scenario, the UAV could transmit its identity certificate to the ground system via UWB, which can then use blockchain technology to verify the authenticity of the certificate. The ground system can access the blockchain ledger to confirm that the UAV meets the necessary requirements, such as having the proper authorization or being up to date with maintenance checks.

This process has potential for significantly improve the security and efficiency of UAV landing systems, as it allows for real-time verification of the UAV's identity and status. The decentralized and secure nature of blockchain can also prevent unauthorized access or tampering of the identity certificates, ensuring that only authorized UAVs are able to land. 

\begin{figure}
    \centering
    \includegraphics[width=\textwidth]{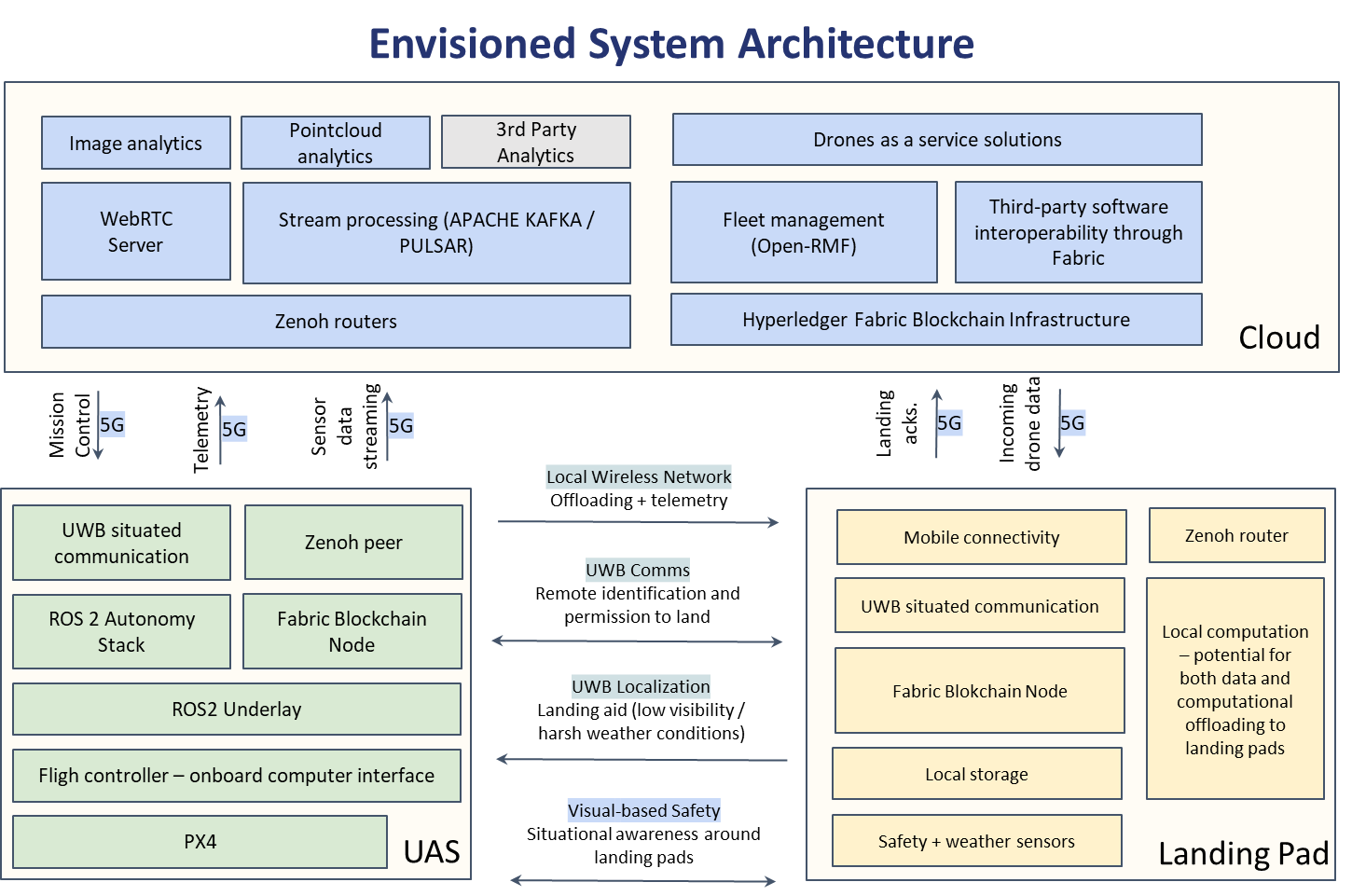}
    \caption{Envisioned system software architecture.}
    \label{fig:architecture}
\end{figure}

In terms of the specific software implementation, the envisioned system architecture is illustrated in Fig.~\ref{fig:architecture}. Only part of such a system is actually implemented within the context of this chapter. However, we envision a system architecture where the key components are the cloud backend systems, the UAS system and the ground infrastructure (e.g., landing pad), which acts as an edge gateway to the cloud. Within this architecture, we show how UWB-based situated communication can be used for both remove identification and proof of location, as well as in aiding in the landing process for more accurate localization. Additionally, this interaction can be extended to, e.g., obtaining permission to dock or take-off. At the cloud end, we see Zenoh as one of the most promising technologies for such distributed systems, allowing for performant and scalable communication and computing, as recent studies demonstrate~\cite{liang2023performance, xianjia2023loosely}. On the fleet management and interfacing side, in addition to Hyperledger Fabric, a version of Open-RMF adapted to UAVs has significant potential as an open-source solution to the problem~\cite{sim2022development, zhang2022distributed}.

\begin{figure}
    \centering
    \includegraphics[width=\textwidth]{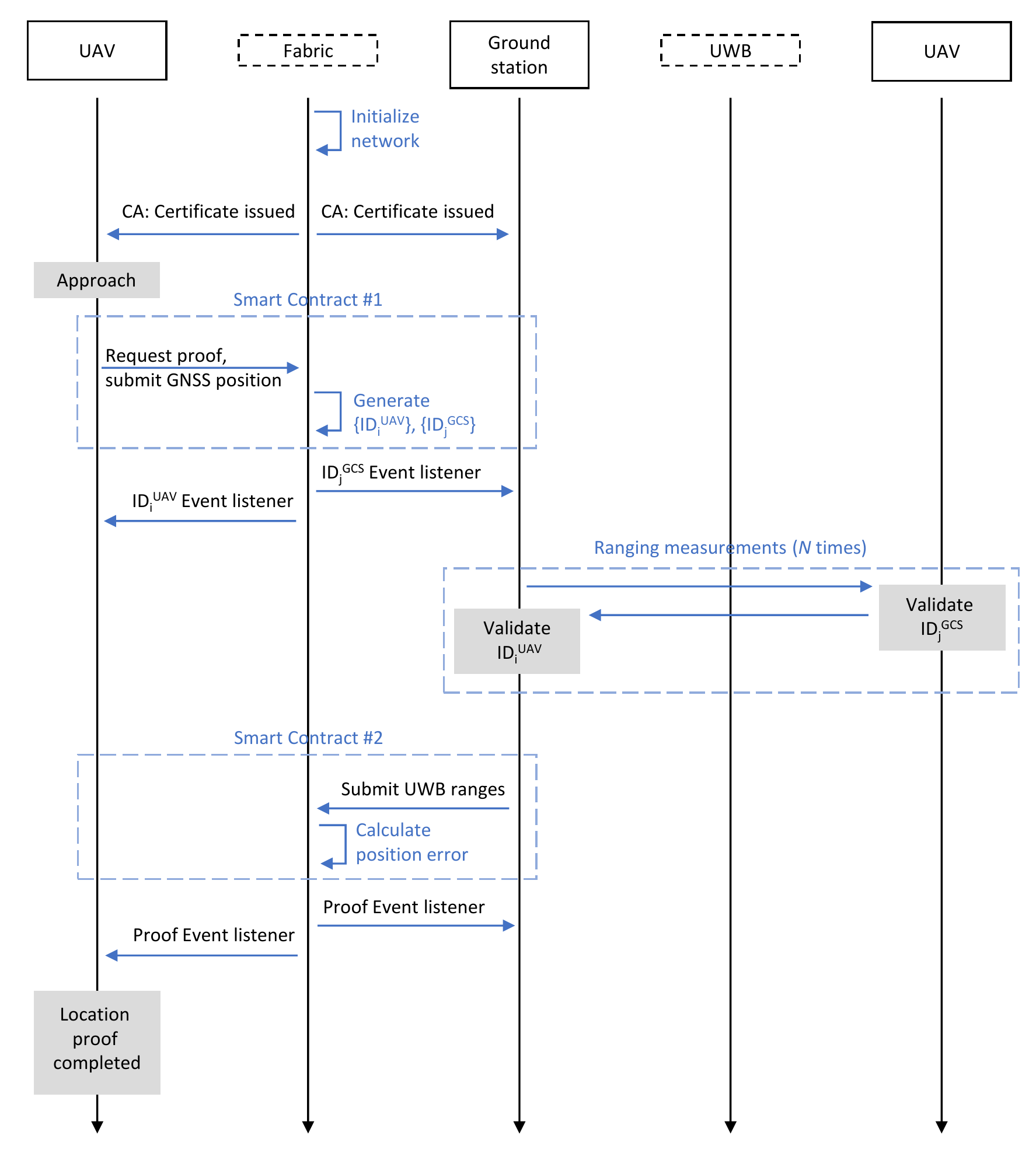}
    \caption{Illustration of the proof of location process. The vertical axis indicates time (flowing down). The UAV is represented on both sides of the figure to differentiate communication between the UAV and the ground platform (GCS) through the Fabric Network (Fabric column) or through the UWB radios (UWB column).}
    \label{fig:process_time}
\end{figure}

The process of proof of location we propose, which is illustrated in Fig.~\ref{fig:process_time}, would operate as follows:

\begin{enumerate}
    \item UAV issues a timestamped notification as a Fabric blockchain transaction and generates a uniquely identifiable code for both UAV and ground platform, to be used for ranging and validating. The UAV then initiates a proof of location procedure.
    \item The ground platform receives the request through blockchain and issues a polling message through the UWB with the received UAV ID and platform ID to potential receivers nearby.
    \item UAV receives the message and compares the unique IDs in the blockchain. If the identities are paired, the UAV replies the polling message from the ground platform.
    \item Platform receives the reply from UWB polling message.
    \item The platform's UWB node initiates two way ranging with the UAV. Additionally it sends its encrypted identity.
    \item UAV receives part of the encrypted information for platform identification. It replies to the polling using UWB if the identity matches.
    \item Platform is able to compute the distance to the UAV for localization. The smart contract compares the GPS location data with the UWB location data. If it matches, it then sends a message to prove the UAV location and verify the authorization for interaction, e.g., landing.
    \item If UAV is authorized, it is given permission to land. 
\end{enumerate}

%% file: sections/04_Methodology.tex
\section{Methods and Experiments}

To demonstrate the viability of the proposed approach, we target a proof of concept demonstration where we focus on implementing the UWB-based PoL approach with Hyperledeger Fabric, which provides enhanced security, reliability and accountability to ensure the UAVs are at specific
locations during critical actions.

\subsection{UWB Ranging}

UWB for localization can be implemented with different ranging modalities, with the one we are using being Time of Flight (ToF). It consists of measuring the time it takes to exchange messages between a pair of UWB devices to obtain their distance, and then triangulating the distances obtained from different pairs to get the position~\cite{torrico2022towards}. With a decentralized system, a minimum of two messages is necessary for the distance calculation and a minimum of four nodes for the position. Leveraging the need to exchange messages for the UWB localization, we utilize these messages to also exchange the UAV and ground platform uniquely identifiable code for authentication. Once an event is triggered in the Hyperledger Fabric and the codes are generated, the platform UWBs will share its codes with the UAV UWBs, by embedding them in one of the messages used for ToF ranging. If the code received in the UAV matches the one obtained from the Fabric, the UAV UWBs respond with their codes for the platform to verify. This provides verification on the side of the UAV and the platform, while also providing the necessary exchange of messages for the relative localization to occur. The system can then compare the GPS location and verify if the UAV computed location matches to authorize the interaction.

\subsection{Hyperledger Fabric Network}

Hyperledger Fabric network functioned as a secure data transport layer that connects different ROS subsystems. We utilize our recent event-driven architecture that utilizes ROS-Fabric application to handle callbacks originating from both the ROS and Fabric networks~\cite{lei2023event}. Whenever new messages are published to specific topics, a ROS event will be generated. In accordance with the specified configuration of the bridge application, such messages can be subsequently relayed to the Fabric network in the form of transactions, which may involve the creation or modification of assets located in distinct channel chaincodes. Fabric channel events are triggered by the confirmation of new transactions on the blockchain, representing an integral component of the system's event-driven architecture.
In our case, both UAV and landing pad act as one host in the Fabric network and requests are transmitted in the blockchain as an event.  

\subsection{Proof of Location}

We implement one chaincode (smart contract) and two Fabric applications for the experimental setup. The smart contract has a simple data structure that can support any ROS message type and minimal functions, including creation, edition, query and deletion of assets. The main idea is to register channel events with transaction types and payloads, which enables the Fabric applications to listen to channel events without prior knowledge of their structure or methods.
The two applications are specifically designed for interactions between the Hyperledger Fabric network and the ROS computational graph of a given system. To manage the assets, each host is equipped with two applications, called \textit{Fabric publisher} and \textit{Fabric subscriber}, with the former responsible for transmitting data from ROS to Fabric and the latter performing the reverse operation. 
In our case, the UAV initiates the procedure by issuing a timestamped notification as a Fabric blockchain transaction and generating a unique code for both the UAV and the landing platform. The platform then receives the request and issues a polling message with IDs to potential receivers nearby. If the identities are paired, the UAV responds to the polling message using UWB, allowing the platform to compute the distance to the UAV for localization in the smart contract. If the location data matches, the platform sends a message to verify the authorization of landing and the proof of location through blockchain.

\subsection{Experimental Setup}

The current implementation relies on the PX4 flight control stack and a companion computer running our autonomy stack through the Robot Operating System (ROS). The UWB ranging is implemented with Decawave DWM1001 transceivers and a custom firmware. The UAV is assembled from a TDrone M690B frame, with a flight time over 50\,min and 2\,Kg of payload. 

\vspace{1em}
\begin{figure}
    \centering
    \includegraphics[width=.95\textwidth]{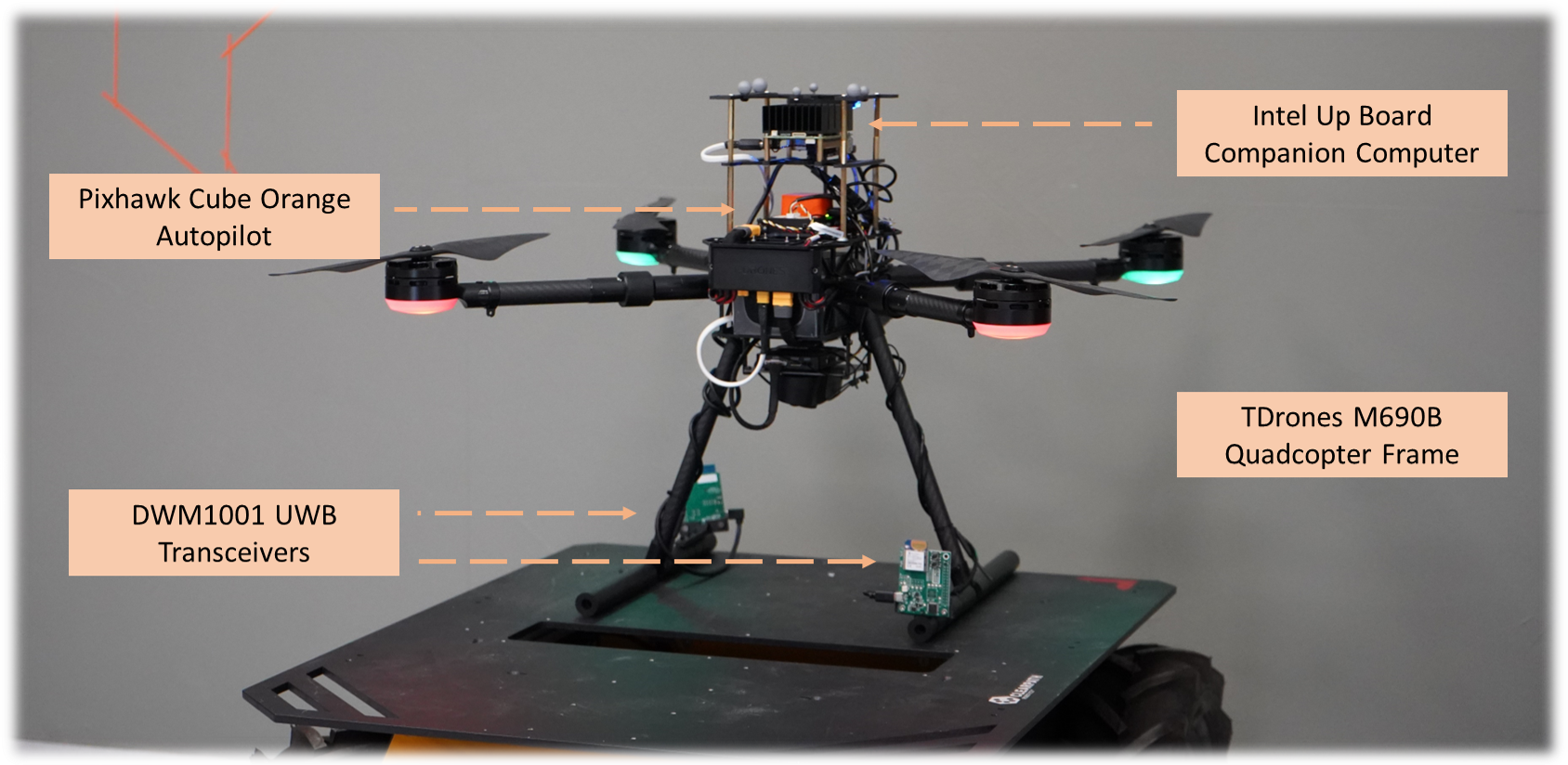}
    \caption{Illustration of the UAV and UWB sensors used in our experiments.}
    \label{fig:UAV_uwb}
\end{figure}

\subsubsection{Hardware.} The Fabric network is set up with two different computers for UAV and landing platform, with various hardware capabilities. The computational load of running the Fabric network remains vastly negligible, as demonstrated in our previous research~\cite{salimi2022towards, lei2023event}. The two hosts have Intel i7-9750H and Intel i3-1215U processors, and 64\,GB and 16\,GB of memory, respectively. We use DWM1001 UWB nodes with a custom firmware. Only one of the UWB nodes is used in our experiment for now. The platform is equipped with four UWB nodes to get a more accurate UAV location data.

\subsubsection{Software.} The Fabric applications are implemented in NodeJS and the chaincode in Golang. Both UAV and landing platform run ROS Noetic under Ubuntu 20.04. Two Fabric applications are then deployed in each host to interface with the ROS systems. UWB nodes are programmed and used to authenticate the identification as well as locate the position of the UAV. 

\subsubsection{Location data.} We use an external motion capture (MOCAP) system with 16 Optitrack cameras to determine the precise location data of the UAV and landing platform, which is employed as the ground truth reference system. 

\begin{figure}
    \centering
    \setlength{\figurewidth}{.8\textwidth}
    \setlength{\figureheight}{.8\textwidth}
    \input{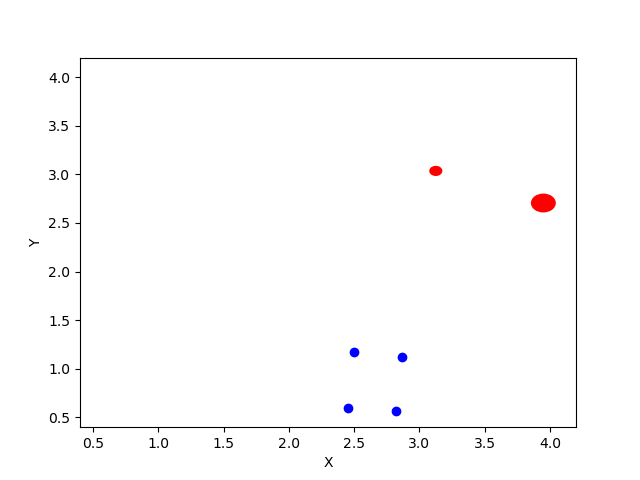}
    \caption{First experiment for remote identification at short distances with two different measurements while the UAV is moving. Secure short-distance location can aid in precision landing systems.}
    \label{fig:test01}
\end{figure}

\begin{figure}
    \centering
    \setlength{\figurewidth}{.8\textwidth}
    \setlength{\figureheight}{.8\textwidth}
    \input{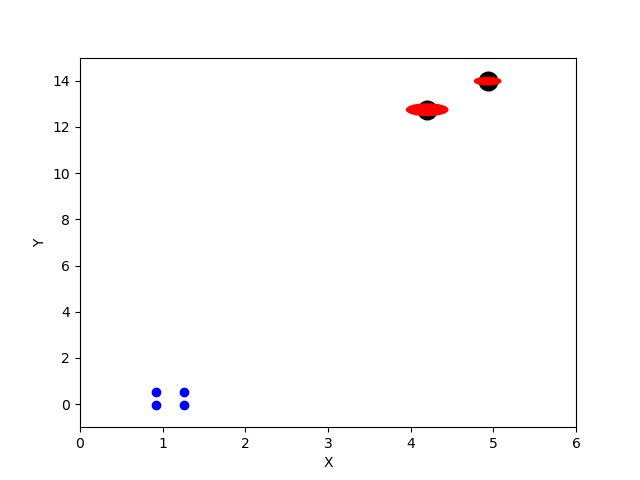}
    \caption{Second experiment for remote identification at larger distance with two different measurements while the UAV is moving.}
    \label{fig:test02}
\end{figure}

\subsection{Experimental Results}

This subsection reports experimental results for the proof of concept based on the  implementation described above. To validate the proof of location system, we perform experiments at both short and long distances. In Fig.~\ref{fig:test01}, we show the positions of the UWB transceivers in the ground station (fixed, in known locations) and the UAV (location to be validated). These positions are obtained from the MOCAP system and emulate GNSS positions in actual outdoor flights. In order to validate the location given by the UAV through the smart contract in the Fabric blockchain, we predefine a maximum error buffer with a 1\,m radios, shown in the figure in yellow.

The results in both Fig.~\ref{fig:test01} and Fig.~\ref{fig:test02} show that the location of the UAV is effectively validated in four different occasions. While the estimated location error obtained from the UWB measurements becomes more comparable to the predefined maximum error buffer in the larger distance experiment, we can argue that a buffer of 1\,m is rather conservative as GNSS errors in single-constellation receivers are easily in the order of meters, specially in urban areas. In any case, the output of the proof of location smart contract can be altered to express instead the error itself, or the likelihood that the UAV is in fact in the position it claims.

%% file: tex/fig1.tex
% This file was created with tikzplotlib v0.10.1.
\begin{tikzpicture}

    \definecolor{darkgray176}{RGB}{176,176,176}
    
    \definecolor{darkgray141160203}{RGB}{141,160,203}
    \definecolor{darkslategray38}{RGB}{38,38,38}
    \definecolor{indianred1967882}{RGB}{196,78,82}
    \definecolor{lavender234234242}{RGB}{234,234,242}
    \definecolor{lightgray204}{RGB}{204,204,204}
    \definecolor{lightslategray129114179}{RGB}{129,114,179}
    \definecolor{mediumseagreen85168104}{RGB}{85,168,104}
    \definecolor{steelblue76114176}{RGB}{76,114,176}
    
    \definecolor{color0}{rgb}{0.1,0.1,0.1}
    \definecolor{color1}{rgb}{0.83921568627451,0.152941176470588,0.156862745098039}
    \definecolor{color2}{rgb}{0.12156862745098,0.466666666666667,0.705882352941177}
    \definecolor{color3}{rgb}{0.580392156862745,0.403921568627451,0.741176470588235}
    \definecolor{color4}{rgb}{0.890196078431372,0.466666666666667,0.76078431372549}

    \begin{axis}[
        width=\figurewidth,
        axis background/.style={fill=white},
        axis line style={white},
        legend cell align={left},
        legend style={
          fill opacity=1,
          draw opacity=1,
          text opacity=1,
          at={(0.75,0.25)},
          anchor=south west,
          draw=lightgray204,
          % legend columns=2,
          % /tikz/every even column/.append style={column sep=0.1cm},
          font=\scriptsize
        %   fill=lavender234234242
        },
        axis equal image,
        tick align=outside,
        x grid style={white!69.0196078431373!black},
        xlabel=\textcolor{darkslategray38}{\(\displaystyle x\) (m)},
        minor tick num = 1,
        minor grid style={dashed},
        xmajorgrids,
        xmajorticks=true,
        xtick style={color=darkslategray38},
        y grid style={white!69.0196078431373!black},
        ylabel=\textcolor{darkslategray38}{\(\displaystyle y\) (m)},
        xminorgrids,
        xminorgrids=true,
        ymajorgrids,
        ymajorticks=true,
        yminorgrids,
        yminorgrids=true,
        ytick style={color=darkslategray38},
        %
        %
        %
        %
        %
        %
        % tick align=outside,
        % tick pos=left,
        % x grid style={darkgray176},
        % xlabel={X},
        xmin=0.4, xmax=5.2,
        % xtick style={color=black},
        % y grid style={darkgray176},
        % ylabel={Y},
        ymin=0.4, ymax=3.5,
        % ytick style={color=black}
    ]
        \draw[draw=red,fill=red] (axis cs:3.95,2.705) circle (0.09);
        \draw[draw=yellow, fill=yellow, fill opacity=0.15] (axis cs:3.95,2.705) circle (1);
        \draw[draw=red,fill=red] (axis cs:3.126,3.035) circle (0.045);
        \draw[draw=yellow, fill=yellow, fill opacity=0.15] (axis cs:3.126,3.035) circle (1);
        \addplot [draw=blue, fill=blue, mark=*, only marks, mark size=1.42]
            table{%
            x  y
            2.5 0.6
            2.5 1.15
            2.85 1.15
            2.85 0.6
        };
        \addlegendentry{Ground UWB transceivers}
        \addplot [draw=black, fill=black, mark=*, only marks, mark size=1.42]
            table{%
            x  y
            3.95 2.705
            3.126 3.035
        };
        \addlegendentry{Broadcasted UAV location}
        \addplot [draw=red, fill=red, mark=*, only marks, mark size=1.42]
            table{%
            x  y
            0 0
        };
        \addlegendentry{Estimated UWB location error}
        \addplot [draw=yellow, fill=yellow, fill opacity=0.15, mark=*, only marks, mark size=1.42]
            table{%
            x  y
            -2 -2
        };
        \addlegendentry{Predefined max. error buffer}
    \end{axis}

\end{tikzpicture}

%% file: tex/fig2.tex
% This file was created with tikzplotlib v0.10.1.
\begin{tikzpicture}

    \definecolor{darkgray176}{RGB}{176,176,176}
    
    \definecolor{darkgray141160203}{RGB}{141,160,203}
    \definecolor{darkslategray38}{RGB}{38,38,38}
    \definecolor{indianred1967882}{RGB}{196,78,82}
    \definecolor{lavender234234242}{RGB}{234,234,242}
    \definecolor{lightgray204}{RGB}{204,204,204}
    \definecolor{lightslategray129114179}{RGB}{129,114,179}
    \definecolor{mediumseagreen85168104}{RGB}{85,168,104}
    \definecolor{steelblue76114176}{RGB}{76,114,176}
    
    \definecolor{color0}{rgb}{0.1,0.1,0.1}
    \definecolor{color1}{rgb}{0.83921568627451,0.152941176470588,0.156862745098039}
    \definecolor{color2}{rgb}{0.12156862745098,0.466666666666667,0.705882352941177}
    \definecolor{color3}{rgb}{0.580392156862745,0.403921568627451,0.741176470588235}
    \definecolor{color4}{rgb}{0.890196078431372,0.466666666666667,0.76078431372549}

    \begin{axis}[
        width=\figurewidth,
        height=\figureheight,
        axis background/.style={fill=white},
        axis line style={white},
        legend cell align={left},
        legend style={
          fill opacity=1,
          draw opacity=1,
          text opacity=1,
          at={(0.75,0.4)},
          anchor=south west,
          draw=lightgray204,
          % legend columns=2,
          % /tikz/every even column/.append style={column sep=0.1cm},
          font=\footnotesize
        %   fill=lavender234234242
        },
        axis equal image,
        tick align=outside,
        x grid style={white!69.0196078431373!black},
        xlabel=\textcolor{darkslategray38}{\(\displaystyle x\) (m)},
        minor tick num = 1,
        minor grid style={dashed},
        xmajorgrids,
        xmajorticks=true,
        xtick style={color=darkslategray38},
        y grid style={white!69.0196078431373!black},
        ylabel=\textcolor{darkslategray38}{\(\displaystyle y\) (m)},
        xminorgrids,
        xminorgrids=true,
        ymajorgrids,
        ymajorticks=true,
        yminorgrids,
        yminorgrids=true,
        ytick style={color=darkslategray38},
        %
        %
        %
        %
        %
        %
        % tick align=outside,
        % tick pos=left,
        % x grid style={darkgray176},
        % xlabel={X},
        xmin=-1.4, xmax=8.2,
        % xtick style={color=black},
        % y grid style={darkgray176},
        % ylabel={Y},
        ymin=-1.4, ymax=15.2,
        % ytick style={color=black}
    ]
        \draw[draw=red,fill=red] (axis cs:4.2,12.745) circle (0.25);
        \draw[draw=yellow,fill=yellow, fill opacity=0.15] (axis cs:4.2,12.745) circle (1);
        \draw[draw=red,fill=red] (axis cs:4.931,13.982) circle (0.16);
        \draw[draw=yellow,fill=yellow, fill opacity=0.15] (axis cs:4.931,13.982) circle (1);
        \addplot [draw=blue, fill=blue, mark=*, only marks, mark size=1.42]
        table{%
            x  y
            1.26 0.518
            1.26 -0.0393
            0.918 -0.0393
            0.918 0.518
        };
        \addlegendentry{Ground UWB transceivers}
        \addplot [draw=black, fill=black, mark=*, only marks, mark size=1.42]
            table{%
            x  y
            4.2 12.745
            4.931 13.982
        };
        \addlegendentry{Broadcasted UAV location}
        \addplot [draw=red, fill=red, mark=*, only marks, mark size=1.42]
            table{%
            x  y
            -2 -2
        };
        \addlegendentry{Estimated UWB location error}
        \addplot [draw=yellow, fill=yellow, fill opacity=0.15, mark=*, only marks, mark size=1.42]
            table{%
            x  y
            -2 -2
        };
        \addlegendentry{Predefined max. error buffer}
        
    \end{axis}
    
\end{tikzpicture}

%% file: sections/05_Conclusion.tex
\section{Conclusion and Future Work}

We have proposed a novel approach to proving the location of aerial robots or autonomous aerial systems through situated communication with ground stations. The method proposed in this paper relies on UWB ranging measurements and a Hyperledger Fabric blockchain network to validate both the identity and the position of a UAV that communicates with fixed ground infrastructure, given that the position of such infrastructure is known a priori. Through a proof of concept implementation, we show that the proposed solution is viable, and we provide sample Fabric smart contracts and applications required for such a system. Our experiments also demonstrate that the UWB-Fabric-based proof of location can be executed in real-time and has potential for scalability.

In future work, we will integrate the proposed solutions with ROS\,2 and the PX4 autopilot project to deliver a more generic and power-on-and-ready solution.